\pdfoutput=1

\documentclass[11pt]{article}

\usepackage{EMNLP2023}
\usepackage{amsmath}

\usepackage[utf8]{inputenc}
\usepackage{times}
\usepackage{latexsym}
\usepackage{graphicx}
\usepackage[T1]{fontenc}
\usepackage[T5]{fontenc}
\usepackage{booktabs}
\usepackage{caption}
\usepackage[vietnamese,english]{babel}
\usepackage[utf8]{inputenc}
\usepackage{amsmath}
\usepackage{amssymb}
\usepackage{multirow}
\usepackage{booktabs}
\usepackage{algorithm}
\usepackage{algpseudocode}
\usepackage{float} 
\usepackage{subcaption} 
\usepackage{setspace}
\usepackage{microtype}
\usepackage[table,xcdraw]{xcolor}
\usepackage{inconsolata}
\usepackage{arydshln} 
\usepackage{hyperref}

\title{Multilingual Generative Retrieval via Cross-lingual Semantic Compression}

\author{
Yuxin Huang$^{1,2}$,
Simeng Wu$^{1,2}$,
Ran Song$^{1,2}$\thanks{Corresponding author.},
Yan Xiang$^{1,2}$, \\
\textbf{Yantuan Xian}$^{1,2}$,
\textbf{Shengxiang Gao}$^{1,2}$,
\textbf{Zhengtao Yu}$^{1,2}$ \\
$^{1}$Faculty of Information Engineering and Automation, \\
Kunming University of Science and Technology, Kunming, China \\
$^{2}$Yunnan Key Laboratory of Artificial Intelligence, Kunming, China \\
\{huangyuxin2004, simengggwu, song$_{-}$ransr\}@163.com, xianyt@kust.edu.cn, \\
sharonxiang@126.com, \{gaoshengxiang.yn, ztyu\}@hotmail.com
}

\begin{document}
\maketitle
\begin{abstract}


Generative Information Retrieval is an emerging retrieval paradigm that exhibits remarkable performance in monolingual scenarios.
However, applying these methods to multilingual retrieval still encounters two primary challenges, cross-lingual identifier misalignment and identifier inflation. 
To address these limitations, we propose Multilingual Generative Retrieval via Cross-lingual Semantic Compression (MGR-CSC), a novel framework that unifies semantically equivalent multilingual keywords into shared atoms to align semantics and compresses the identifier space, and we propose a dynamic multi-step constrained decoding strategy during retrieval. 
MGR-CSC improves cross-lingual alignment by assigning consistent identifiers and enhances decoding efficiency by reducing redundancy. 
Experiments demonstrate that MGR-CSC achieves outstanding retrieval accuracy, improving by $6.83\%$ on mMarco100k and $4.77\%$ on mNQ320k, while reducing document identifiers length by $74.51\%$ and $78.2\%$, respectively.
We publicly release our dataset and code at \url{https://github.com/simengggg/MGR-CSC}

\end{abstract}

\section{Introduction}

Multilingual Information Retrieval (MIR) serves as a critical component in natural language processing,  particularly in applications such as cross-border e-commerce~\cite{li2020e-commerce} and cross-lingual search systems~\cite{xu2021artificial}.
The core need lies in developing models that can effectively process multilingual queries and retrieve relevant documents across different languages~\cite{zhang2019improving,dwivedi2016survey}. 
Traditional translation-based approaches compromise retrieval quality due to error propagation in machine translation pipelines~\cite{chandra2017assessing}.
Recently multilingual pre-trained language models (PLMs)~\cite{xue2020mt5,conneau2019xlmr} demonstrate improved performance by encoding cross-lingual content into joint semantic spaces~\cite{yarmohammadi2019robust}. 
However, precise cross-lingual alignment and end-to-end semantic matching remain significant challenges for pervious methods.

\begin{figure}[t!]
    \centering
    \includegraphics[width=\linewidth]{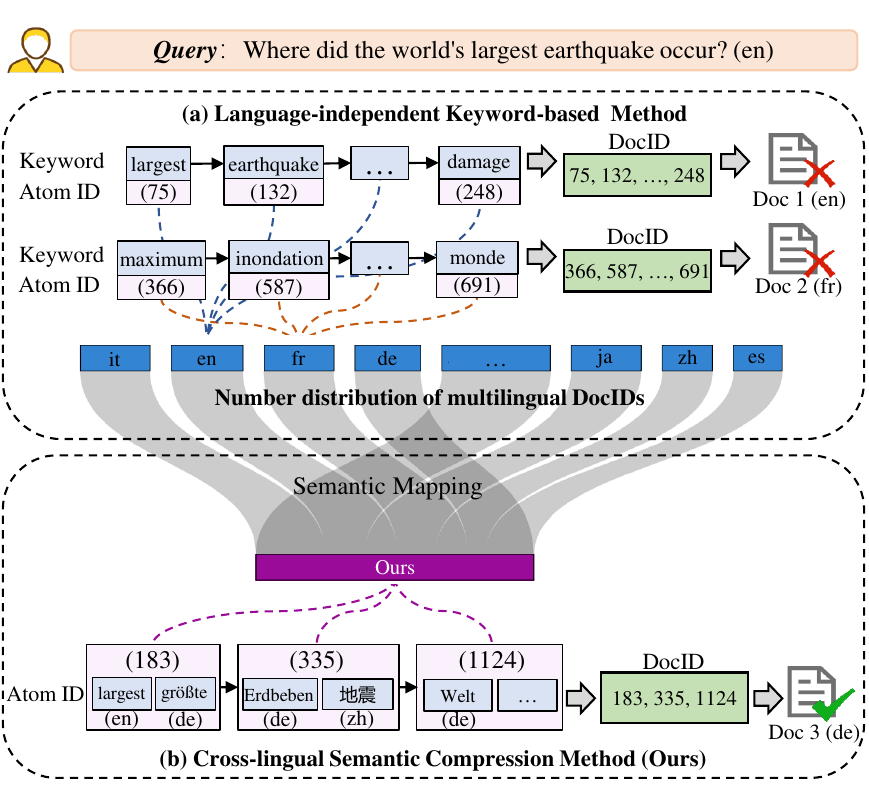} 
    \caption{(a) illustrates the language-independent keyword approach, where the model biases toward query-language DocIDs during decoding. In contrast, (b) demonstrates MGR-CSC successfully retrieves target documents via semantic clustering, enabling more reliable identification.}
    \label{fig1}
\end{figure}


Generative Information Retrieval (GIR) offers a paradigm shift by leveraging the model's parametric memory to store documents and directly generates document identifiers (DocIDs)~\cite{tay2022dsi,zhuang2022dsi-qg, sun2023genret}.
This approach leverages generative PLMs to learn and encode direct associations between documents and their unique DocIDs.
Unlike previous paradigms, GIR offers an alternative end-to-end framework by utilizing generative PLMs to directly map queries to relevant DocIDs. 
This generative approach inherently addresses the mentioned challenges by learning direct associations between multilingual queries and DocIDs.


However, the application of GIR to multilingual scenarios confronts two fundamental limitations:
(i) \textbf{Cross-lingual Identifier Misalignment.} Existing DocIDs are constructed using language-independent encoding schemes, creating isolated semantic mapping spaces for each language. 
As shown in Figure~\ref{fig1} (a), queries in a specific language lead the model to preferentially generate DocIDs for documents in the same language, such as English and French. 
This phenomenon is a direct result of cross-lingual identifier misalignment, which consequently hinders the transfer of multilingual knowledge through shared latent document-level representations.
(ii) \textbf{Multilingual Identifier Inflation.} The number of existing DocIDs increases manifold as the number of languages grows in multilingual GIR.
As shown in Figure~\ref{fig1} (a), keyword-based methods assign different atom IDs to semantically equivalent keywords across languages. 
For instance, the English word \textit{largest} might be assigned atom ID \textit{75}, while its French counterpart \textit{maximum} is assigned atom ID \textit{366}, despite their semantic equivalence. 
The combinatorial expansion of unique DocIDs in multilingual documents intensifies this challenge, posing significant hurdles for memory efficiency and runtime performance in autoregressive decoding architectures.
Therefore, a semantically consistent DocID is essential for cross-lingual alignment and efficient decoding.


In this paper, we present \textbf{M}ultilingual \textbf{G}enerative \textbf{R}etrieval via  \textbf{C}ross-lingual \textbf{S}emantic \textbf{C}ompression (MGR-CSC), a novel framework that unifies multilingual keyword semantics into shared atom IDs, assigns DocIDs to documents, and employs multi-step constrained decoding.
A high-level overview of MGR-CSC is shown in Figure~\ref{fig1} (b). MGR-CSC performs cross-lingual semantic compression by mapping semantically equivalent multilingual keywords to shared atom IDs. 
For example, both the English word \textit{largest} and its German counterpart \textit{größte} are mapped to atom ID \textit{183}. In addition, it applies dynamic decoding constraints to guide generation.
Specifically, MGR-CSC contain three key parts. 
First, we extract explicit keywords from each multilingual document, and the document is represented by a set of multiple keywords.
Secondly, these keywords are projected into a shared latent space using unsupervised clustering. 
Within this latent space, semantically equivalent expressions are assigned the same atom ID, which effectively compresses the multilingual identifier space.
Finally, we introduce a multi-step dynamic constraint decoding strategy. 
The initial decoding step leverages the global frequency distribution of atom IDs to guide selection. 
And a refinement step narrows the selection space based on constraints between atomic IDs from preceding steps.
Comprehensive experiments on multiple benchmark datasets show that MGR-CSC achieves outstanding performance, surpassing existing multilingual generative retrieval approaches by ${6.83\%}$ on mMarco100k and ${4.77\%}$ on mNQ320k. 
Furthermore, it substantially reduces the number of DocID tokens by $74.51\%$ on mMarco100k and $78.2\%$ on mNQ320k.

The contributions of this paper are as follows:

\begin{itemize}
\item We propose DocID construction approach for multilingual documents in MGR-CSC, enabling alignment and compact representation across languages.

\item We propose a dynamic constrained multi-step decoding framework in MGR-CSC to reduce decoding complexity.

\item Our experiments on multilingual benchmarks demonstrate the method's effectiveness and generalization in cross-lingual retrieval.
\end{itemize}

\section{Related Work}
\subsection{Multilingual Information Retrieval}
Multilingual information retrieval (MIR) seeks to semantically align queries and documents across languages, enabling cross‐lingual access. 
Early MIR relied on translation, such as queries~\cite{elayeb2018querytrans1,chandra2017assessing}, documents~\cite{yarmohammadi2019robust}, or both~\cite{dwivedi2016survey} to balance efficiency and accuracy in practical systems. 
The advent of multilingual pretrained language models such as mBERT~\cite{devlin2019bert} and XLM‑R~\cite{conneau2019xlmr} shifted attention to vector‑based retrieval. 
In this paradigm, queries and documents are embedded into a shared space for similarity comparison~\cite{yu2020study,zhang2022hike}, thus eliminating external translation~\cite{oard2008trans-ir} and improving understanding in low‑resource languages.
Nonetheless, these approaches are limited by a fixed encode–match–rank pipeline lacking end‐to‐end optimization~\cite{tay2022dsi,sun2023genret} and by contrastive learning’s dependence on scarce parallel data~\cite{karpukhin2020dense}.

\subsection{Generative Information Retrieval}
Generative approaches are applied in fields such as information retrieval \cite{tay2022dsi,sun2023genret} and knowledge graphs\cite{song2024multilingual,abu2024knowledge}.
By capitalizing on the memorization capabilities of pre-trained language models\cite{song2024does}, GIR directly generates DocIDs during inference, facilitating an end-to-end retrieval process. 
Current DocID representations primarily fall into two categories: atomic DocIDs and string DocIDs. 
For atomic DocIDs, Tay et al. \cite{tay2022dsi} proposed constructing DocIDs using randomly assigned identifiers or cluster-based embedding layers.
In contrast, string DocIDs employ semantically meaningful strings, such as document title~\cite{tang2023se-dsi} or keywords, such as TSGen~\cite{zhang2024tsgen}, Novo~\cite{wang2023novo}. 
Although, current GIR are primarily designed for monolingual settings, making effective adaptation to multilingual contexts difficult to achieve. 
To bridge this gap, we propose a semantic compression approach unifying cross-lingual lexical representations into a shared semantic space for multilingual GIR.

\section{Methodology}
In this section, we elaborate on our proposed method, Multilingual Generative Retrieval via Cross-lingual Semantic Compression, termed \textbf{MGR-CSC}. 
The core idea is to assign an unique DocID to each document by leveraging semantically similar key information across documents in different languages. 
This shared representation enables effective cross-lingual alignment and facilitates end-to-end semantic matching within a unified retrieval framework.

\begin{figure*}[t!]
    \centering
    \includegraphics[width=\textwidth]{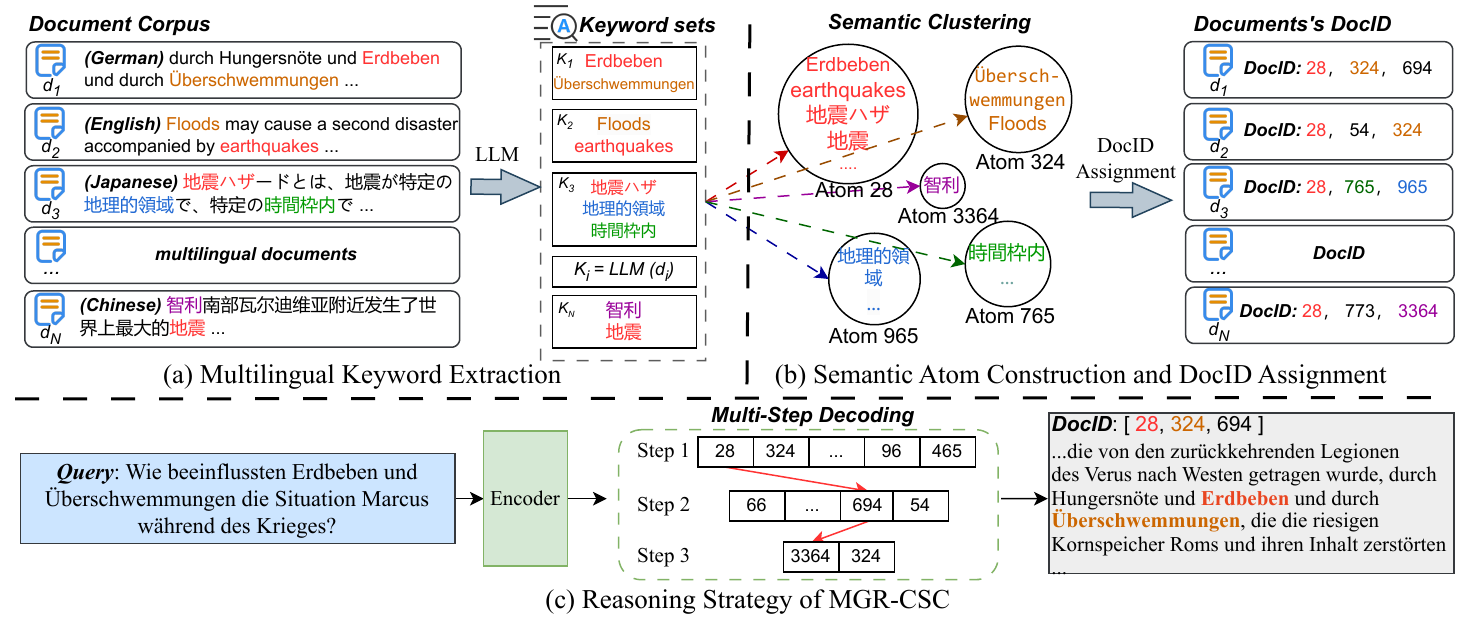} 
    \caption{An overview of MGR-CSC's DocID allocation and reasoning. (a) shows how to extract keywords of documents; (b) shows clustering is performed based on cross-lingual semantic similarity of keywords, with each cluster represented by an atom and each document assigned a unique DocID; (c) shows MGR-CSC's reasoning strategy, which returns the identifier corresponding to the query, and narrows the decoding range at each step.}
    \label{fig2}
\end{figure*}

As Figure~\ref{fig2} illustrates, MGR-CSC consists of three components: (1) extracting distinctive keywords from multilingual documents, (2) clustering multiple keywords into one semantic atom, and assigning each document a unique DocID as a sequence of atoms, (3) during retrieval, the model generates the DocID atom-by-atom under dynamic decoding constraints.

\subsection{Multilingual Keyword Extraction}
\label{section3.2}
To capture the essential semantics of multilingual documents, we extract a fixed set of $m$ keywords from each document using a prompt-based Large Language Model (LLM). 
These keywords serve as a compact representation of the multilingual document's content. 

Formally, given a document \( d_i \), we extract a fixed number \( m \) of keywords as follows:
\begin{equation}
K_i = \{k_i^1, k_i^2, \ldots, k_i^m\} = \text{LLM}(d_i),
\end{equation}
where \( K_i\) is treated as a semantically compact representation of the document \( d_i \).
And \( k_{i}^{m} \) denotes the \( m \)-th keyword of document \( i \). 
We utilize a standardized extraction process across all language documents in order  to ensure the consistency of semantic representations.

\subsection{Semantic Atom Construction and DocID Assignment}

Based on the extracted multilingual keywords, we construct language-independent semantic  atoms and assign unique DocIDs to documents. 

First, we aggregate all keywords from the entire document collection into a single global set $K$. 
Although a specific keyword may appear in the keyword sets $K_i$ of multiple documents, it is represented only once as a distinct element in the global set $K$. 
Formally, the global keyword set $K$ is defined as the union of the individual document keyword sets $K_i$ for all $d_i \in \mathcal{D}=[d_1,\ldots,d_N]$:
\begin{equation}
K = \bigcup_{i=1}^N K_i,
\end{equation}
where set $K$ contains $n$ unique keywords, and $n = |K|$ is the total number of distinct keywords in the collection. 
Let $\{\hat{k}_1, \hat{k}_2, \ldots, \hat{k}_n\}$ denote the set of unique keywords. 
Each keyword $\hat{k}_i$ is encoded into a dense vector representation $v_i \in \mathbb{R}^d$ through a pre-trained text encoder. We construct a similarity matrix $S \in \mathbb{R}^{n \times n}$ where each entry $S_{ij} = \frac{v_i \cdot v_j}{\|v_i\| \|v_j\|}$ represents the cosine similarity between keywords $\hat{k}_i$ and $\hat{k}_j$. 

Keywords are clustered together when their pairwise similarity meets or exceeds a predefined threshold $\theta \in [0,1]$.
Specifically, a fixed number of cluster centers is chosen. 
Keywords with similarity above the threshold $\theta$ are directly assigned to the nearest cluster center, while the others form single clusters.
As a result, the process gives $C$ clusters with $C < N$, since similar keywords are grouped together.
As the keywords within each cluster are semantically similar, each cluster is represented by an atom, thereby obtaining a set of atoms \(\mathcal{A} = \{a_1, a_2, \ldots, a_c\}\), where $a_c$ is the atomic representation of the $c$-th cluster.

Finally, since each keyword $\hat{k}_i$ belongs to a cluster represented by an atom $a$, it can consequently be represented by $a$. 
Based on section~\ref{section3.2}, we obtain the set of keywords $K_i$ for each document $d_i$. Each keyword $k_i^m$ is converted to its atom representation. 
Let $a_{k_i^m}$ denote the atom representing the keyword $k_i^m$. 
The resulting sequence of atom representations is used as the unique DocID for each document $d_i$,
\begin{equation}
\text{DocID}_i = [a_{k_i^1}, a_{k_i^2}, \ldots, a_{k_i^m}]= [a_1, a_2, \ldots, a_m],
\end{equation}
where $\text{DocID}_i$ represents the DocID of the i-th document. 
This ensures that all multilingual documents are assigned DocID representations with consistent length and shared semantic space.

\subsection{Dynamic Constrained Multi-Step Decoding}
In prior methods~\cite{zhuang2022dsi-qg, tang2023se-dsi}, after obtaining document representations with unique DocIDs, the decoding process requires selecting from the complete set of $N$ documents at each step. 
For a decoding sequence of length $m$, the method produces a search space scaling as $O(N^m)$.
Our proposed method, transforms the retrieval process into a multi-step decoding process of length $m$ within a constrained space comprising $c$ semantic atoms, thereby effectively compressing the decoding space to $O(C^m)$.

To handle the output space, we employ a dynamic constrained multi-step decoding mechanism. 
During the retrieval process, based on the query $q$, the model generates the DocID of the target document $d_q$ through the following approach,
\begin{equation}
P(\mathrm{DocID} \mid q) = \prod_{t=1}^m P(a_t \mid a_{<t}, q),
\end{equation}
where $P$ represents the generation probability of the DocID. 

The retrieval process generates the target document's DocID through a multi-step decoding procedure under dynamic constraints. 
At each decoding step \( t \) (\( 1 \leq t \leq m \)), the model predicts the \( t \)-th atom \( a_t \) based on the query \( q \) and the previously generated prefix \( \text{DocID}_{<t} = [a_1, \ldots, a_{t-1}] \). 
The candidate atom set \( \mathcal{A}_t \) is defined as:  
\begin{equation}  
\mathcal{A}_t = \left\{ a_{k_i^t} \, \middle| \, k_i^t \in \text{Constraint}(K_i) \right\},  
\end{equation}  
where \( \text{Constraint}(K_i) \) represents the range of documents available under the prefix constraint, and the prefix \( \text{DocID}_{<t} = [a_1,a_2,\ldots,a_{t-1}] \). 
The optimal atom at step \( t \) is selected via:  
\begin{equation}  
a_t = \underset{a_{k_i^t} \in \mathcal{A}_t}{\arg\max} \, P(a_{k_i^t} \mid a_1, \ldots, a_{t-1}, q),  
\end{equation}  
with \( \text{DocID}[t] = a_t \). Through \( m \) iterations, the complete DocID is obtained as:  
\begin{equation}  
\text{DocID} = [a_1, a_2, \ldots, a_m].  
\end{equation}

Algorithm \ref{alg:multistage} outlines the decoding process applying this dynamic constraint.

\begin{algorithm}[!t]
\caption{Dynamic Constrained Multi-Step Decoding}
\label{alg:multistage}
\begin{algorithmic}[1]
\small
\State \textbf{Input} Query $q$, Keyword set $K$, Documents $\mathcal{D} = \{d_1, \ldots, d_N\}$, Atom set $\mathcal{A} = [a_1,\ldots,a_c]$
\State \textbf{Output} Target document DocID
\vspace{0.5em}
\State Initialize empty DocID sequence: $\text{DocID} \leftarrow [\,]$
\vspace{0.5em}
\For{$t = 1$ \textbf{to} $m$}
  \If{$t = 1$}
    \State $\text{prefix} \leftarrow \text{DocID}[\,]$
  \Else
    \State $\text{prefix} \leftarrow \text{DocID}[1:t-1]$
    \EndIf
    \vspace{0.5em}
    \State $\mathcal{A}_t \leftarrow \{a_{k_i^t} \mid k_i^t \in \text{Constraint}(K_i)\}$
    \vspace{0.5em}
  \State Compute decoding distribution:
  \State \quad $P_t \leftarrow \{P(a_{k_i^t} \mid \text{prefix}, q) \mid a_{k_i^t} \in \mathcal{A}_t\}$
  
  \State Select optimal atom:
  \State \quad $a_t \leftarrow \arg\max(P_t)$
  
  \State Update DocID: $\text{DocID}[t] \leftarrow a_t$
\EndFor
\vspace{0.5em}
\State \textbf{Return} $\text{DocID} = [a_1, \ldots, a_m]$
\end{algorithmic}
\end{algorithm}

\section{Experiment}

\subsection{Datasets}
To comprehensively evaluate the model's performance in multilingual document retrieval, we conduct comparative analyses leveraging both standard public benchmarks and our newly constructed multilingual dataset. 
Below we provide overview of these evaluation datasets:

\textbf{mMarco100K}~\footnote{https://github.com/unicamp-dl/mMARCO} is a multilingual retrieval benchmark constructed through neural machine translation of the original English MS MARCO dataset~\cite{bonifacio2021mmarco}, covering more than 30 languages, consists of a document and a question-answer pair. 
We randomly sampled non-parallel corpus data in 7 languages, with about 15k data in each language, and a total of about 100k data. 
Among them, the data set is divided into 6.5k data as a validation set and the rest as a training set.

\textbf{mNQ320K} is a novel multilingual retrieval dataset developed in this study to overcome the limitations of existing resources in low and medium resource language scenarios. 
It consists of query and document pairs, including about 307K training data and 8K verification data, constructed through a systematic translation methodology following the mMARCO framework. 
Specifically, We create an extended version of the NQ320K~\cite{kwiatkowski-etal-2019-natural} dataset by translating the original data into seven medium-resource languages spanning diverse language families: Afrikaans (af), French (fr), Arabic (ar), Hindi (hi), Macedonian (mk), Swedish (sv), and Vietnamese (vi).

\begin{table*}[!t] 
\centering
\small
\setlength{\tabcolsep}{2.7pt}
\renewcommand{\arraystretch}{1.3} 
\begin{tabular}{cccccccccccccccccc}
\hline
\textbf{\rotatebox{90}{ }} & \multirow{2}{*}{\textbf{Method}} 
& \multicolumn{2}{c}{en $\Rightarrow$ 
 oth} & \multicolumn{2}{c}{fr $\Rightarrow$ oth} & \multicolumn{2}{c}{de $\Rightarrow$ oth} 
& \multicolumn{2}{c}{it $\Rightarrow$ oth} & \multicolumn{2}{c}{es $\Rightarrow$ oth} & \multicolumn{2}{c}{ja $\Rightarrow$ oth} 
& \multicolumn{2}{c}{zh $\Rightarrow$ oth} & \multicolumn{2}{c}{AVG} \\
& & @1 & @10 & @1 & @10 & @1 & @10 & @1 & @10 & @1 & @10 & @1 & @10 & @1 & @10 & @1 & @10 \\
\hline
\multirow{8}{*}{\rotatebox{90}{\textbf{mMarco 100k}}} 
& BM25 & 17.53 & 35.34 & 11.45 & 26.48 & 6.50 & 15.93 & 10.48 & 22.85 & 14.01 & 28.87 & 0.00 & 0.00 & 0.58 & 1.17 & 8.65 & 18.66 \\ 
& Colbert-xm & 62.11 & 83.45 & 45.45 & 70.66 & 36.62 & 58.76 & 38.09 & 61.12 & 36.00 & 64.11 & 41.19 & 69.26 & 39.41 & 65.63 & 42.69 & 67.57 \\
& mColbert & 52.42 & 73.17 & 40.71 & 68.17 & 39.98 & 63.45 & 38.55 & 61.84 & 35.05 & 63.23 & 36.49 & 62.38 & 33.33 & 58.55 & 39.50 & 64.40 \\
& LaBSE & 58.04 & 79.97 & 51.10 & 76.71 & 42.29 & 66.71 & 46.45 & 70.67 & 48.03 & 74.92 & 39.22 & 65.02 & 40.73 & 70.22 & 46.55 & 72.03 \\ 
& DSI & 10.96 & 23.52 & 10.01 & 23.83 & 9.16 & 21.34 & 9.20 & 21.84 & 9.58 & 22.01 & 9.43 & 20.50 & 8.75 & 20.54 & 9.50 & 21.80 \\ 
& DSI-QG & 74.72 & 87.67 & 66.05 & 83.46 & 61.58 & 79.96 & 65.02 & 82.09 & 66.90 & 83.94 & 62.09 & 80.76 & 60.77 & 78.46 & 65.21 & 82.34 \\ 
& SE-DSI & \textbf{76.57} & 87.05 & \textbf{71.48} & 83.58 & 67.15 & 79.40 & 68.06 & 82.18 & 70.62 & 84.23 & 66.35 & 79.91 & 67.03 & 80.87 & 69.49 & 82.43 \\ 
& Ours & 72.07 & \textbf{91.27} & 68.53 & \textbf{87.74} & \textbf{70.73} & \textbf{90.79} & \textbf{69.37} & \textbf{89.01} & \textbf{71.74} & \textbf{91.60} & \textbf{68.68} & \textbf{88.32} & \textbf{67.37} & \textbf{86.13} & \textbf{69.78} & \textbf{89.26} \\ 
\hline
\textbf{\rotatebox{90}{ }} & \multirow{2}{*}{\textbf{Method}} 
& \multicolumn{2}{c}{af $\Rightarrow$ oth} & \multicolumn{2}{c}{fr $\Rightarrow$ oth} & \multicolumn{2}{c}{ar $\Rightarrow$ oth} 
& \multicolumn{2}{c}{hi $\Rightarrow$ oth} & \multicolumn{2}{c}{mk $\Rightarrow$ oth} & \multicolumn{2}{c}{sv $\Rightarrow$ oth} 
& \multicolumn{2}{c}{vi $\Rightarrow$ oth} & \multicolumn{2}{c}{AVG} \\
& & @1 & @10 & @1 & @10 & @1 & @10 & @1 & @10 & @1 & @10 & @1 & @10 & @1 & @10 & @1 & @10 \\
\hline
\multirow{8}{*}{\rotatebox{90}{\textbf{mNQ 320k}}}
& BM25 & 11.73 & 22.01 & 12.41 & 25.45 & 11.49 & 24.98 & 10.75 & 23.13 & 11.39 & 22.48 & 13.41 & 26.94 & 11.84 & 25.41 & 11.77 & 24.16 \\  
& Colbert-xm & 13.01 & 28.19 & 17.26 & 35.47 & 12.48 & 22.75 & 14.64 & 29.99 & 13.73 & 29.20 & 20.49 & 37.00 & 17.36 & 35.55 & 15.57 & 31.16 \\
& mColbert & 17.79 & 36.78 & 18.86 & 38.67 & 14.30 & 27.80 & 14.05 & 27.49 & 15.27 & 30.35 & 22.68 & 43.01 & 18.82 & 36.52 & 17.39 & 34.37 \\
& LaBSE & 22.14 & 45.13 & 21.96 & 46.25 & 14.40 & 31.40 & 20.47 & 42.48 & 21.44 & 42.78 & 25.18 & 51.65 & 24.13 & 46.32 & 21.29 & 43.29 \\ 
& DSI & 0.11 & 0.46 & 0.10 & 0.41 & 0.00 & 0.00 & 0.00 & 0.20 & 0.10 & 0.40 & 0.00 & 0.34 & 0.10 & 0.60 & 0.05 & 0.33 \\ 
& DSI-QG & 24.34 & 45.03 & \textbf{25.08} & 48.50 & \textbf{20.95} & 39.48 & 16.15 & 37.34 & 19.84 & 42.17 & 25.90 & 52.03 & 20.76 & 42.22 & 21.32 & 43.05 \\ 
& SE-DSI & 14.17 & 26.51 & 20.62 & 36.06 & 15.42 & 27.06 & 15.84 & 28.05 & 15.35 & 28.32 & 22.52 & 36.49 & 17.27 & 33.33 & 16.76 & 29.96 \\ 
& Ours & \textbf{26.74} & \textbf{49.71} & 24.97 & \textbf{50.47} & 20.31 & \textbf{44.39} & \textbf{21.04} & \textbf{42.57} & \textbf{23.58} & \textbf{48.76} & \textbf{28.80} & \textbf{52.88} & \textbf{23.38} & \textbf{47.66} & \textbf{24.11} & \textbf{48.06} \\ 
\hline
\end{tabular}
\caption{Performance at Recall@1 and Recall@10 on the mMARCO100K and mNQ320K under cross-lingual retrieval settings. 
Bolded values indicate the best performance among all comparison methods.}
\label{table:main}
\end{table*}

\subsection{Baselines}
We conduct comparisons with both traditional retrieval approaches and recent multilingual generative retrieval models. 
To ensure fairness, we reproduce known advanced generative retrieval methods capable of handling multilingual retrieval.

\textbf{BM25} \cite{robertson2009bm25} represents a standard sparse retrieval model that leverages inverted index structures and operates based on exact keyword correspondence.


\textbf{LaBSE}~\cite{feng-etal-2022labse} a multilingual sentence encoder that supports 109 languages and maps text from different languages into a unified vector space for cross-lingual retrieval tasks.

\textbf{mColBERT}~\cite{khattab2020colbert} a multilingual version of ColBERT that employs late interaction mechanisms for dense retrieval.

\textbf{ColBERT-xm}~\cite{louis2024colbertxm} a cross-lingual interaction-based retrieval model that enhances multilingual retrieval through fine-grained token-level interactions.

\textbf{DSI} \cite{tay2022dsi} a generative retrieval method that treats documents as training input and constructs DocIDs using hierarchical clustering-based approach.

\textbf{DSI-QG} \cite{zhuang2022dsi-qg} a generative retrieval method that trains a query generation model, using short queries to represent the original documents and random numbers to represent DocIDs.

\textbf{SE-DSI} \cite{tang2023se-dsi} a generative retrieval method that utilizes strings containing semantic information as DocIDs. 
In this study, the titles of multilingual documents are used as the DocIDs for this method.

\subsection{Detailed Implementation}

In our experiments, all methods were reproduced on the mT5-base model~\footnote{https://huggingface.co/google/mt5-base} based on the Transformer architecture. 
Following the work of previous researchers~\cite{zhuang2022dsi-qg}, our training data adheres to the approach of representing documents with multilingual queries. 
Specifically, the model used for all pseudo-query generation tasks is Llama3.1-8B~\footnote{https://huggingface.co/meta-llama/Llama-3.1-8B}~\cite{grattafiori2024llama} with a temperature of 0.7, each document sample is converted into 10 multilingual pseudo-queries through model generation. 
For keyword generation, the model employed is Llama3.1-8B with a temperature of 0. 
The model used for semantic similarity calculation is paraphrase-multilingual-MiniLM-L12-v2~\footnote{https://huggingface.co/sentence-transformers}~\cite{reimers-2019-sentence-bert}. 

\textbf{Training} The training was implemented with PyTorch~\cite{paszke2019pytorch} and Transformers~\cite{vaswani2017transformer}.
For mMarco100k, we used a learning rate of $2 \times 10^{-4}$, batch size 128, 50 epochs, and $m{=}3$ keywords;
for mNQ320k, the learning rate was $5 \times 10^{-4}$, with the same batch size and $m$, trained for 100 epochs.
The cross-entropy loss function is employed as the objective function.
All experiments ran on eight NVIDIA A40 GPUs with 46GB.

\textbf{Evaluation Metric} Aligning with previous studies, we evaluate our model's performance on the validation sets of both datasets, employing Recall@1 and Recall@10 as evaluation metrics.  
These metrics indicate the fraction of relevant documents retrieved within the top 1 and top 10 positions. 
Due to the multilingual composition of the candidate document corpus, we present retrieval results for each query language individually.

\begin{figure}[!t] 
    \centering
    \includegraphics[width=\columnwidth]{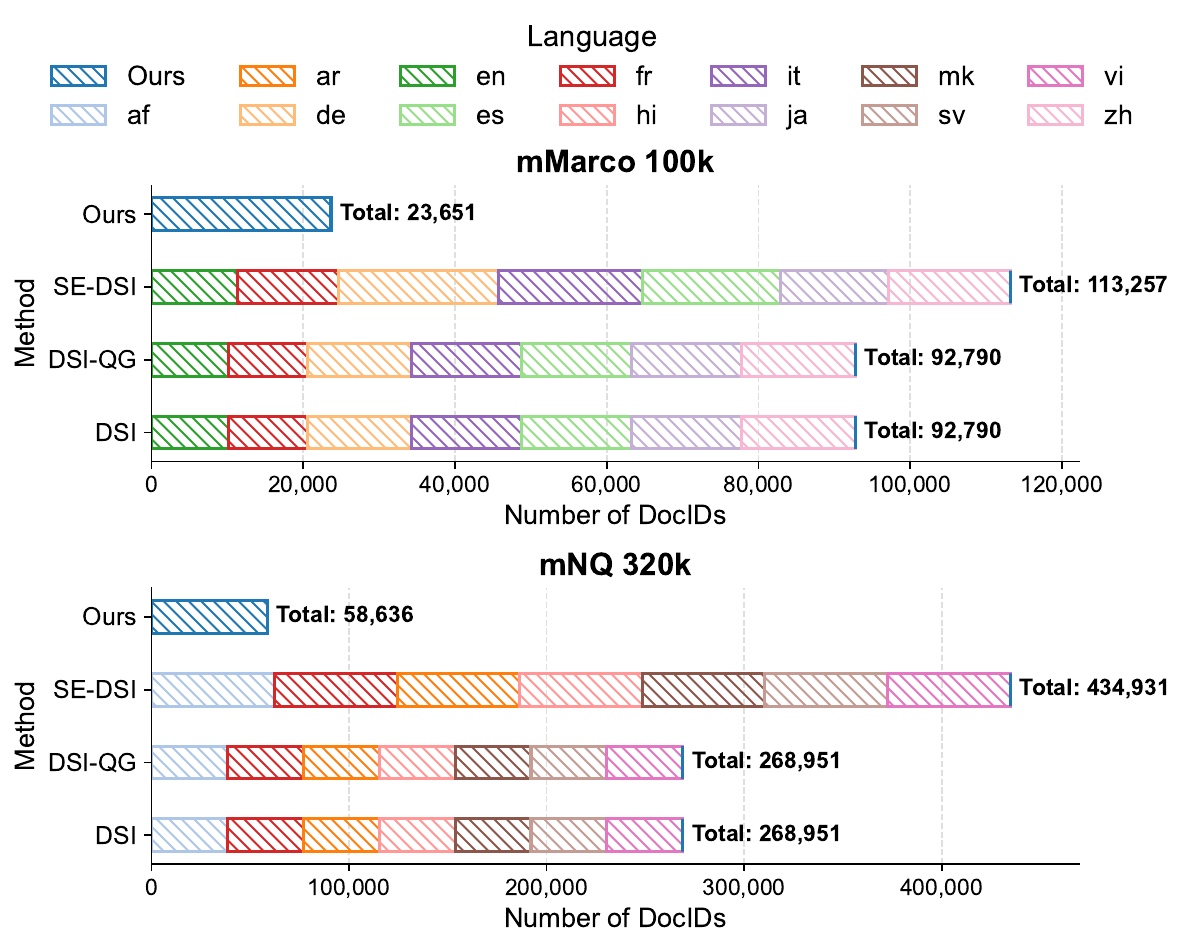}
    \caption{Distribution of the number of DocIDs across different languages for various methods on the mMarco100k and mNQ320k datasets. For SE-DSI, DSI-QG, and DSI, the stacked bar segments represent the distribution of retrieved DocIDs across languages. The overall DocID count for each method is indicated to the right of the corresponding bar.}
    \label{fig:mMarco_docid_usage}
\end{figure}
\subsection{Result on mMarco100k and mNQ320k}
To demonstrate the performance of our model, we conduct a comparative analysis with existing methods. 
Table~\ref{table:main} shows retrieval results on two multilingual benchmarks. 
The mMarco100k dataset, sparse methods such as BM25 perform poorly across languages tasks, especially on non-Latin scripts such as Japanese and Chinese. 
This underscores the constraints of lexical matching in languages with morphological and script diversity.

Dense retrieval methods such as mDPR are more stable but less effective at capturing cross-lingual semantics than generative approaches.
DSI and DSI-QG improve Recall@10 on English but under-perform on syntactically varied languages, reflecting limited generalization.

MGR-CSC demonstrates consistent performance across all tested languages on mMarco100k dataset. 
While its Recall@1 on English and French is slightly below SE-DSI, it achieves the highest Recall@10 in nearly all languages, including German, Italian, Spanish, Japanese, and Chinese. 
This suggests better semantic coverage and decoding stability in multilingual contexts. 

The mNQ320k dataset, which includes a broader range of languages, many with fewer resources, offers a more challenging setting. 
BM25 and DSI exhibit limited retrieval ability in these scenarios. 
mDPR maintains moderate performance but shows sensitivity to linguistic variability.
DSI-QG and SE-DSI improve cross-lingual recall balance over prior methods but exhibit resource-dependent performance. 
SE-DSI shows accelerated degradation under low-resource conditions, particularly in linguistically diverse environments.
\begin{figure}[!t] 
    \centering
    \includegraphics[width=\columnwidth]{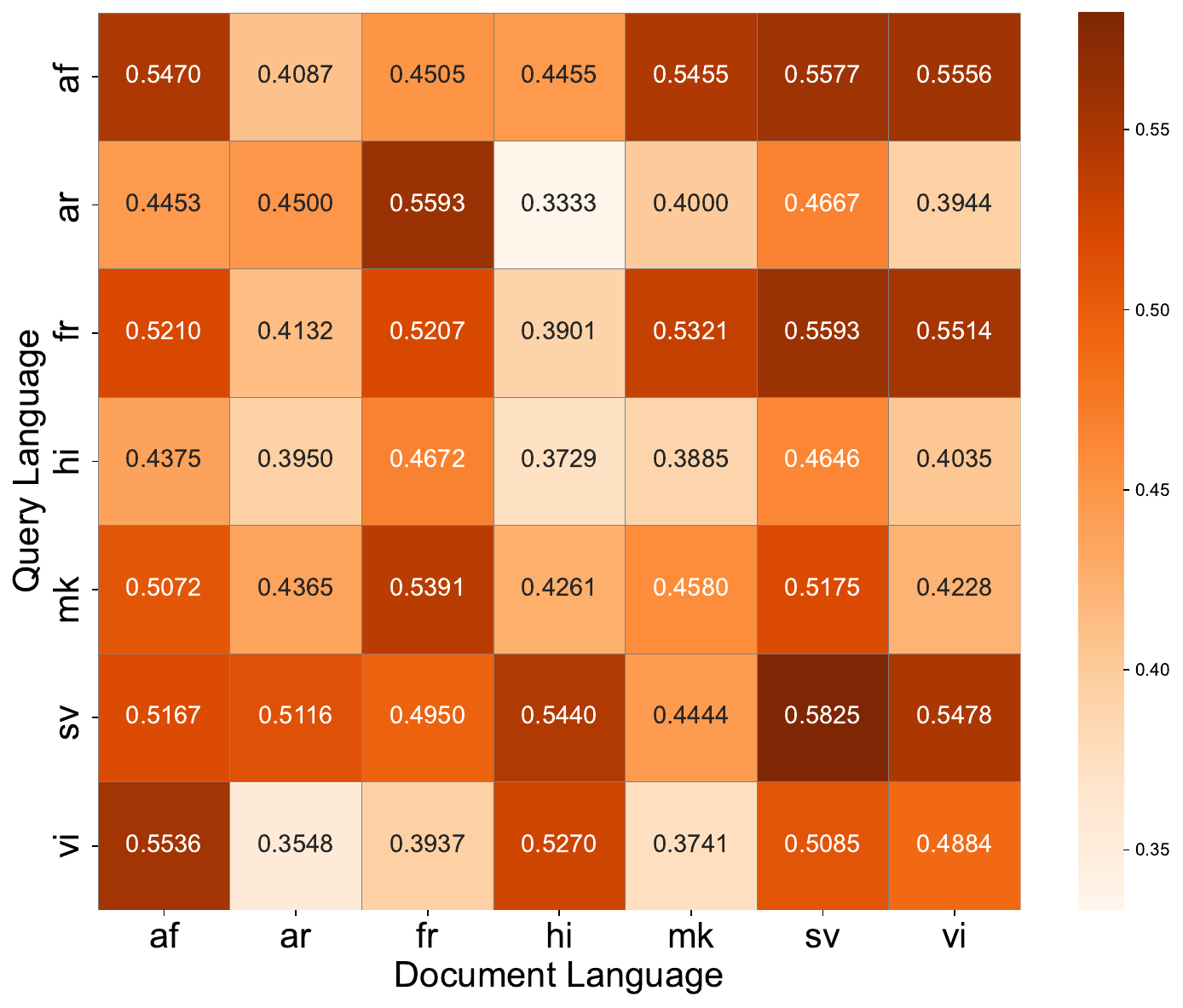}
    \caption{Recall@10 performance of target-language document retrieval with varying source query languages in the mNQ320k dataset}
    \label{fig:language}
\end{figure}

In contrast, MGR-CSC yields stable and high recall on all languages in the mNQ320k dataset, including significant improvements in low and mid resource languages such as Arabic, Hindi, and Macedonian. 
These findings are further supported by the results presented in figure~\ref{fig:language}, which shows the Recall@10 results for the all target language documents across different query languages. 
The experimental results indicate that MGR-CSC successfully maintains retrieval consistency when processing linguistically diverse data and cross-domain scenarios. 
Furthermore, the framework demonstrates strong generalizability across both typologically similar and distinct language families.

\subsection{Quantitative Analysis of DocID Usage}  
To quantitatively compare the decoding range between our proposed methodology and existing GIR approaches, we conducted a comprehensive analysis. 
As depicted in Figure~\ref{fig:mMarco_docid_usage}, under identical dataset conditions, our method consistently achieves the most restricted decoding space. 
This significant reduction in the output range substantially enhances decoding efficiency and scalability, rendering our approach particularly advantageous for large-scale document retrieval.

Furthermore, this reduced decoding space is crucial for addressing challenges in multilingual GIR settings. 
For instance, a single word in high-resource languages, such as English, French, which is typically encoded using one or two tokens. 
In contrast, in low-resource languages, words are commonly represented as a sequence of subword units.
This phenomenon leads to an expanded decoding space and an increased number of decoding steps in low-resource scenarios.
To effectively mitigate this, our methodology introduces atomic integer IDs designed to represent clusters of semantically similar keywords. 
By compressing lexical-level variations into a semantically aligned ID space, this approach effectively minimizes the decoding range and ensures more consistent decoding behavior across linguistic boundaries.

\subsection{Ablation Study} 
To illustrate the contribution of different components, we conducted ablation studies on semantic compression and the decoding strategy separately using the mNQ320k dataset. 
\begin{table}[!h]
    \centering
    \begin{tabular}{ccc}
    \hline
        \multicolumn{1}{l}{Method} & R@1 & R@10 \\ \hline
        \multicolumn{1}{l}{MGR-CSC}  & \textbf{24.11} & \textbf{48.06} \\
        \multicolumn{1}{l}{w/o decoding strategy}  & 13.50 & 31.80 \\ 
        \multicolumn{1}{l}{w/o semantic compression}  & 15.58 & 38.76 \\  \hline
    \end{tabular}
    \caption{Performance at Recall@1 and Recall@10 on the mNQ320K under different ablation settings.}
\end{table}

Experimental results (measured by AVG scores) show that removing either component leads to a significant decline in performance. Specifically, removing the decoding strategy leads to a $10.61\%$ drop in Recall@1 and a $16.26\%$ drop in Recall@10.
Similarly, removing semantic compression results in $8.53\%$ drop in Recall@1 and $9.30\%$ drop in Recall@10.
These results confirm that both semantic compression and the decoding strategy play crucial roles in enhancing retrieval effectiveness.

\subsection{The performance of the number of keywords}

\begin{figure}[!t] 
    \centering
    \includegraphics[width=\columnwidth]{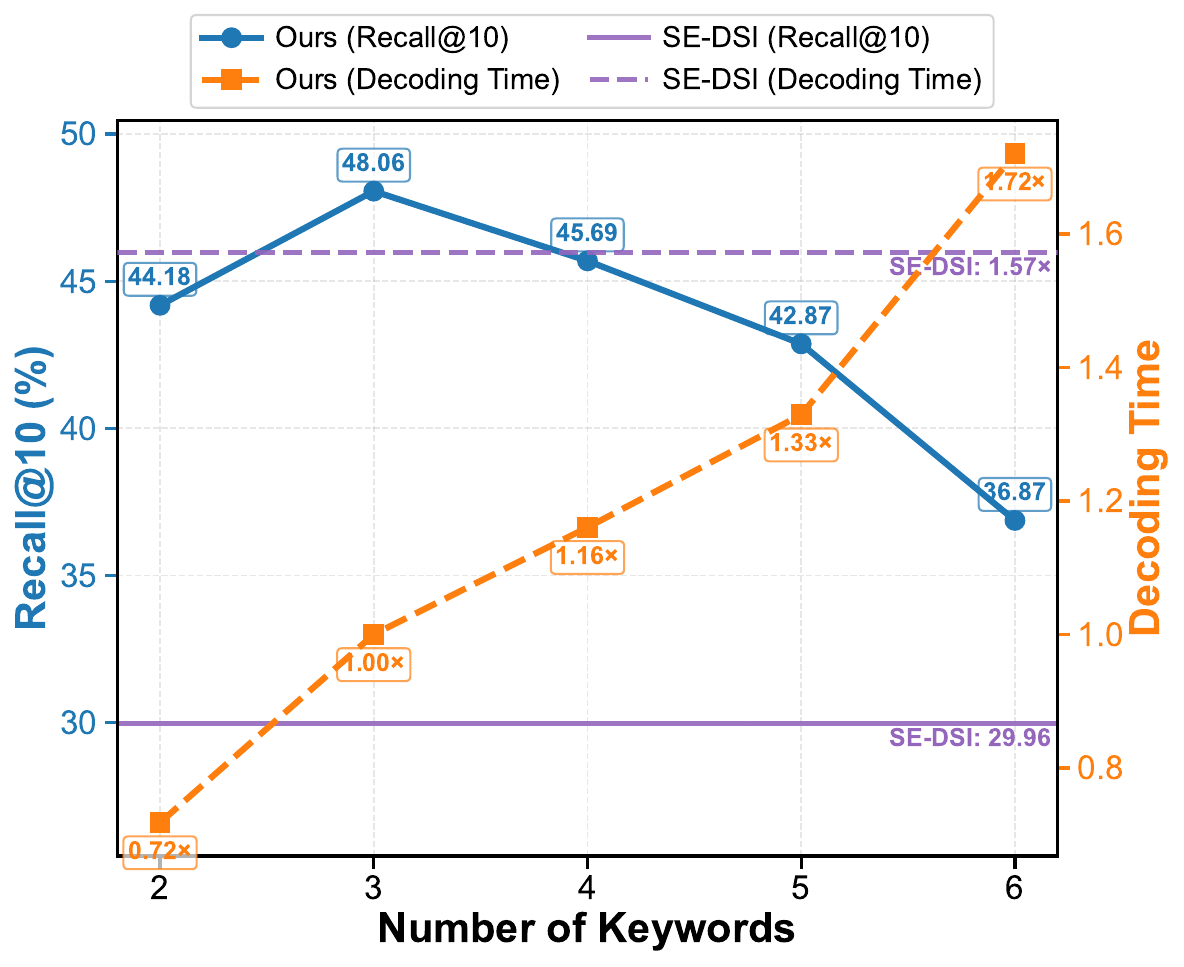}
    \caption{Performance in Recall@10 and decoding time under varying keyword quantities $m$.}
    \label{fig:subfig-number}
\end{figure}
To examine how the keyword number $m$ affects retrieval performance and decoding time, we experimented on the mNQ320k dataset. 
As depicted in Figure~\ref{fig:subfig-number}, increasing the number of keywords enhances semantic representation. 
However, this also increases the length of DocIDs, which consequently slows down decoding and impairs overall performance. 
The best Recall@10 obtained was $48.06\%$ with three keywords. 
Fewer keywords, such as two, restrict semantic coverage. 
In contrast, more keywords, for instance five or six, decrease performance due to increased decoding complexity.

Decoding time scales with DocID length. 
As the number of keywords increases from two to six, the decoding time approximately doubles. 
For comparison, SE-DSI achieves a Recall@10 of $29.96\%$. 
Its latency is comparable to that of our method when employing the longest DocIDs using six keywords. 
This similarity emphasizes the strong correlation between sequence length and inference time.

\subsection{The performance of the semantic similarity threshold}

\begin{figure}[!h]
    \centering
    \includegraphics[width=\columnwidth]{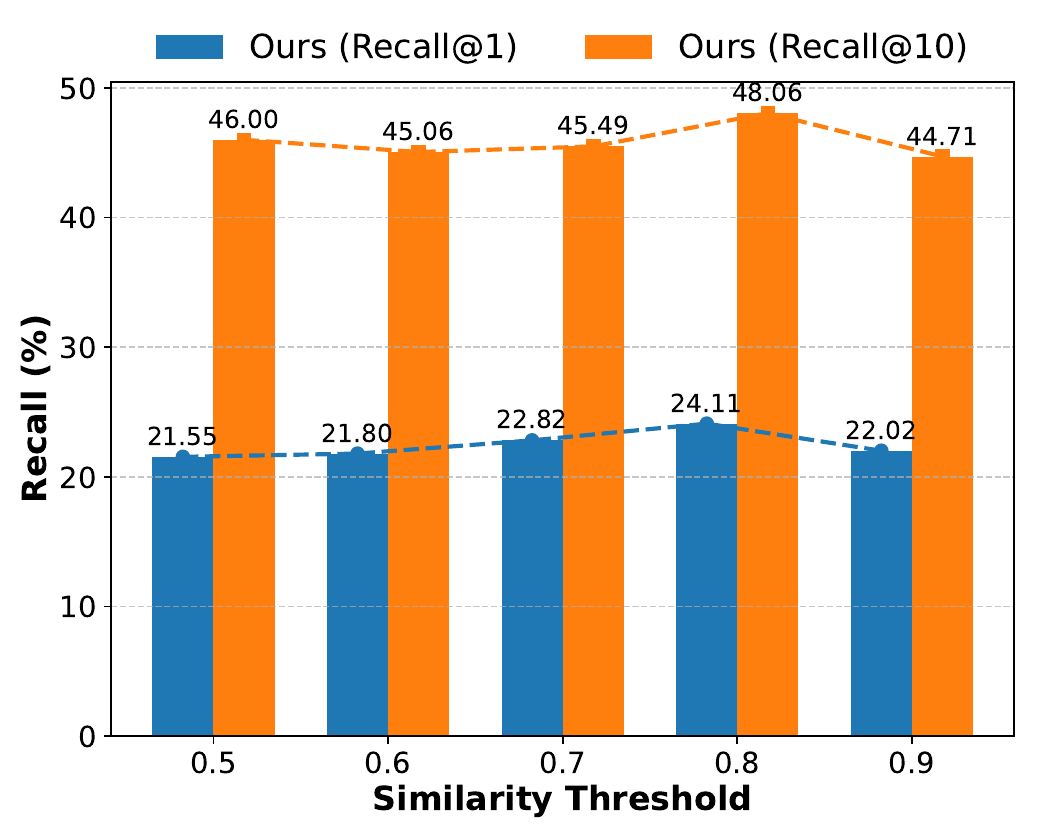}
    \caption{Recall@1 and Recall@10 performance under varying semantic similarity thresholds $\theta$.}
    \label{fig:similarity_recall}
\end{figure}

To assess the sensitivity of keyword clustering to semantic similarity thresholds $\theta$, we evaluated retrieval performance on the mNQ320k dataset with thresholds from 0.5 to 0.9.
As illustrated in Figure~\ref{fig:similarity_recall}, increasing the threshold from 0.5 to 0.8 resulted in a steady improvement in both Recall@1 and Recall@10, indicating that finer-grained clusters enhance retrieval precision.

The best results were achieved at 0.8, with Recall@1 of $24.48\%$ and Recall@10 of $47.89\%$.
However, raising the threshold to 0.9 caused a performance drop, with Recall@1 decreasing by $2.46\%$ to $22.02\%$ and Recall@10 decreasing by $3.18\%$ to $44.71\%$. 
This suggests that an excessively high threshold leads to overly fine-grained clustering, splitting semantically related keywords into separate groups. This reduces semantic generalization and limits the model's retrieval coverage.

\subsection{Case Study}

To demonstrate the effectiveness of our proposed multilingual generative retrieval method, we present a case study on Swedish-Vietnamese cross-lingual retrieval.

\begin{table}[!h]
\centering
\renewcommand{\arraystretch}{1.2}
\small 
\begin{tabular}{p{1.50cm}p{5.6cm}}  
\toprule
\textbf{Type} & \multicolumn{1}{c}{\textbf{Document}} \\
\midrule
\textbf{Theme} & (vi) Danh sách các thủ đô của Úc 

(List of Australian capital cities) \\
\textbf{Content} & (vi) \colorbox{magenta!30}{Úc} có tám thành phố, mỗi thành phố đóng vai trò là trụ sở chính quyền của một tiểu bang hoặc vùng lãnh thổ. Úc được thành lập vào năm 1901. \colorbox{gray!30}{Năm 1927}, trụ sở chính quyền quốc gia đã được di dời và chuyển đến thành phố mới, nơi vẫn tiếp tục đóng vai trò là \colorbox{cyan!30}{thủ đô} quốc gia cho đến ngày nay. Mỗi thủ đô đều có chức năng tư pháp, hành chính và hành chính. ... \\
\textbf{Keywords} & \colorbox{magenta!30}{Úc}, \colorbox{cyan!30}{thủ đô}, \colorbox{gray!30}{Năm 1927}

(Australia, capital, 1927) \\
\textbf{Atom set} & \textcolor{magenta}{46}, \textcolor{cyan}{1788}, \textcolor{gray}{14920} \\
\hline
\hline
\textbf{Query} & (sv) Vad är Australiens huvudstad?  \\
\hdashline
\textbf{DSI-QG} & DocID: 92980 (\textcolor{red}{×}) \\
\textbf{SE-DSI} & DocID: (fr) Territoire de la capitale australienne  (\textcolor{red}{×}) \\
\hdashline
\textbf{Ours-step1} & DocID: 46  \\
\textbf{Ours-step2} & DocID: 46, 1788 \\
\textbf{Ours-step3} & DocID: 46, 1788, 14920 \\
\hline
\textbf{Output} & 46, 1788, 14920 \\
\bottomrule
\end{tabular}
\caption{Case study on mNQ320k.A Vietnamese document is represented by clustered keyword atom sets. For a given Swedish query, the DocID undergoes a step-wise semantic decoding process along semantic dimensions to retrieve the target document.}
\end{table}

Given the Swedish query \textit{``Vad är Australiens huvudstad?''} (What is the capital of Australia?), the model performs multi-step decoding based on clustered semantic atoms. 
Each decoding stage progressively narrows the candidate space by focusing on specific semantic dimensions, first detecting the country entity, then identifying the question type, and finally locating the target concept.

The final DocID is composed of shared semantic atoms, enabling successful retrieval of the corresponding Vietnamese document: \textit{``Danh sách các thủ đô của Úc''} (List of Australian capital cities).

\section{Conclusion}

This paper introduces MGR-CSC, a multilingual generative retrieval method leveraging cross-lingual semantic compression. 
This method employs semantic clustering to reduce multilingual DocIDs and narrow the decoding space, and applies multi-step constrained decoding to restrict DocID generation.
The experimental reveals that our method consistently exhibit outstanding performance compared to existing retrieval approaches when applied to multilingual datasets.

\section*{Limitations}
Since our model is based on multilingual PLM, its multilingual document understanding capability is consequently limited by the capabilities of this base model. 
This limitation is particularly pronounced in the context of low-resource languages.

Furthermore, PLMs are mainly designed for text processing. 
Our existing framework has limited capacity for multimodal information requiring integration of diverse data modalities. 
It can process text components, yet lacks the inherent ability to understand or reason about cross-modality relationships, thereby restricting its performance in complex multimodal scenarios.

\section*{Ethics Statement}
This paper proposes a multilingual generative information retrieval method and conducts experiments on both public datasets and extended datasets. Therefore, there are no data privacy implications in this scenario.

\bibliography{anthology,custom}

\appendix
\clearpage

\onecolumn
\section{Appendix}

\subsection{The results of Recall@100 on the mNQ320k dataset}
\label{Appendix}
\begin{table}[!h]
    \centering
    \begin{tabular}{ccccccccc}
    \hline
        \textbf{Method} & af $\Rightarrow$ oth & fr $\Rightarrow$ oth & ar$\Rightarrow$ oth & hi$\Rightarrow$ oth & mk$\Rightarrow$ oth & sv$\Rightarrow$ oth & vi$\Rightarrow$ oth & AVG \\ \hline
        BM25 & 34.22 & 38.64 & 37.19 & 35.99 & 35.48 & 40.34 & 38.55 & 37.20 \\ 
        Colbert-xm & 43.20 & 52.80 & 35.40 & 48.10 & 48.48 & 54.77 & 50.25 & 47.57 \\ 
        mColbert & 52.58 & 56.15 & 44.56 & 42.09 & 46.44 & 58.48 & 52.60 & 50.41 \\ 
        LaBSE & 65.51 & \textbf{67.48} & 51.35 & \textbf{64.13} & 65.51 & \textbf{70.47} & \textbf{65.88} & 64.33 \\ 
        DSI-QG & 63.04 & 66.39 & 57.84 & 60.56 & 64.43 & 67.07 & 61.98 & 63.04 \\ 
        SE-DSI & 38.45 & 49.70 & 39.84 & 40.35 & 43.84 & 50.61 & 44.43 & 43.88 \\ 
        Ours & \textbf{67.49} & 67.11 & \textbf{63.12} & 62.79 & 65.72 & 69.23 & 64.14 & \textbf{65.66} \\ \hline
    \end{tabular}
\caption{Performance at Recall@100 on the mNQ320K. 
Bolded values indicate the best performance among all comparison methods.}
\label{table4:recall100}
\end{table}

To demonstrate the performance of our method on high-ranks@100, we extend the recall@100 results on the mNQ320k dataset. 
Table~\ref{table4:recall100} presents the results, showing that our method outperforms others in most languages as well as in the average score, with the AVG exceeding the current baseline by $1.33\%$.

\end{document}